# Cross-Target Stance Classification with Self-Attention Networks

**Chang Xu, Cécile Paris, Surya Nepal,** and **Ross Sparks**
CSIRO Data61, Marsfield, NSW, Australia
{Chang.Xu, Cecile.Paris, Surya.Nepal, Ross.Sparks}@data61.csiro.au

## Abstract

In stance classification, the target on which the stance is made defines the boundary of the task, and a classifier is usually trained for prediction on the same target. In this work, we explore the potential for generalizing classifiers between *different* targets, and propose a neural model that can apply what has been learned from a source target to a destination target. We show that our model can find useful information shared between relevant targets which improves generalization in certain scenarios.

## 1 Introduction

Stance classification is the task of automatically identifying users' positions about a specific target from text (Mohammad et al., 2017). Table 1 shows an example of this task, where the stance of the sentence is recognized as favorable on the target *climate change is concern*. Traditionally, this task is approached by learning a target-specific classifier that is trained for prediction on the same target of interest (Hasan and Ng, 2013; Mohammad et al., 2016; Ebrahimi et al., 2016). This implies that a new classifier has to be built from scratch on a well-prepared set of ground-truth data whenever predictions are needed for an unseen target.

An alternative to this approach is to conduct a *cross-target* classification, where the classifier is adapted from different but *related* targets (Augenstein et al., 2016), which allows benefiting from the knowledge of existing targets. For example, in our project we are interested in online users' stances on the approvals of particular mining projects in the country. It might be useful to start with a classifier that is adapted from a related target such as *climate change is concern* (presumably available and annotated), as in both cases users could discuss the impacts from the targets to some common issues, such as the environment or communities.

| **Sentence:** We need to protect our islands and stop the destruction of coral reef. | |
|---|---|
| **Target:** Climate Change is Concern | **Stance:** Favor |

Table 1: An example of stance classification task.

Cross-target stance classification is a more challenging task simply because the language models may not be compatible between different targets. However, for some targets that can be recognized as being related to the same and more general domains, it could be possible to generalize through certain aspects of the domains that reflect users' major concerns. For example, from the following sentence, whose stance is against the approval of a mining project, *"Environmentalists warn the $16 billion coal facility will damage the Great Barrier Reef"*, it can be seen that both this sentence and the one in Table 1 mention the same aspect "reef destruction/damage", which is closely related to the "environment" domain.

In this paper, we focus on cross-target stance classification and explore the limits of generalizing models between different but *domain-related* targets[1]. The basic idea is to learn a set of domain-specific aspects from a source target, and then apply them to prediction on a destination target. To this end, we propose CrossNet, a novel neural model that implements the above idea based on the self-attention mechanism. Our preliminary analysis shows that the proposed model can find useful domain-specific information from a stance-bearing sentence and that the classification performance is improved in certain domains.

---

[1] In this work, the source target is chosen based on common sense. Exploring more sophisticated source target selection methods will be our future work.

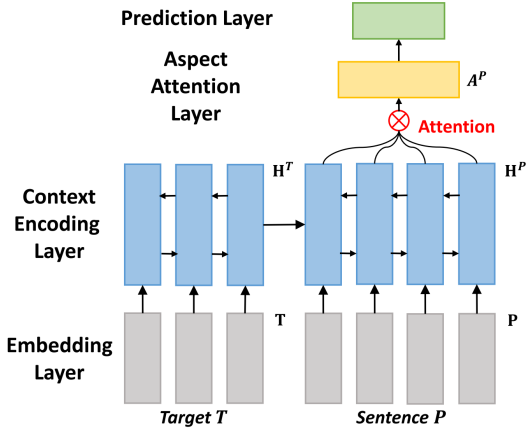

Figure 1: The Architecture of CrossNet.

## 2 Model

In this section, we introduce the proposed model, CrossNet, for cross-target stance classification. Figure 1 shows the architecture of CrossNet. It consists of four layers from the Embedding Layer (bottom) to the Prediction Layer (top). It works by taking a stance-bearing sentence and a target as input and yielding the predicted stance label as output. In the following, we present the implementation of each layer in CrossNet.

### 2.1 Embedding Layer

There are two inputs in CrossNet: a stance-bearing sentence $P$ and a descriptive target $T$ (e.g, *climate change is concern* in Table 1). We use word embeddings (Mikolov et al., 2013) to represent each word in the input as a dense vector. The output of this layer are two sequences of vectors $P = \{p_1, ..., p_{|P|}\}$ and $T = \{t_1, ..., t_{|T|}\}$, where $p, t$ are word vectors.

### 2.2 Context Encoding Layer

In this layer, we encode the contextual information in the input sentence and target. We use a bi-directional Long Short-Term Memory Network (BiLSTM) (Hochreiter and Schmidhuber, 1997) to capture the left and right contexts of each word in the input. Moreover, to account for the impact of the target on stance inference, we borrow the idea of conditional encoding (Augenstein et al., 2016) to model the dependency of the sentence on the target. Formally, we first use a BiLSTM$_T$ to encode the target:

$$[\overrightarrow{h}_i^T \; \overrightarrow{c}_i^T] = \overrightarrow{\text{LSTM}}_T(t_i, \overrightarrow{h}_{i-1}^T, \overrightarrow{c}_{i-1}^T)$$
$$[\overleftarrow{h}_i^T \; \overleftarrow{c}_i^T] = \overleftarrow{\text{LSTM}}_T(t_i, \overleftarrow{h}_{i+1}^T, \overleftarrow{c}_{i+1}^T) \quad (1)$$

where $h \in \mathbb{R}^h$ and $c \in \mathbb{R}^h$ are the hidden state and cell state of LSTM. The symbol $\rightarrow(\leftarrow)$ indicates the forward (backward) pass. $t_i$ is the input word vector at time step $i$.

Then, we learn a conditional encoding of the sentence $P$, by initializing BiLSTM$_P$ (a different BiLSTM) with the final states of BiLSTM$_T$:

$$[\overrightarrow{h}_1^P \; \overrightarrow{c}_1^P] = \overrightarrow{\text{LSTM}}_P(p_1, \overrightarrow{h}_{|T|}^T, \overrightarrow{c}_{|T|}^T)$$
$$[\overleftarrow{h}_{|P|}^P \; \overleftarrow{c}_{|P|}^P] = \overleftarrow{\text{LSTM}}_P(p_{|P|}, \overleftarrow{h}_1^T, \overleftarrow{c}_1^T) \quad (2)$$

It can be seen that the initialization is done by aligning the forward (backward) pass of the two BiLSTMs. The output is a contextually-encoded sequence, $H^P = \{h_1^P, ..., h_{|P|}^P\}$, where $h = [\overrightarrow{h}; \overleftarrow{h}] \in \mathbb{R}^{2h}$ with $[;]$ as the vector concatenation operation.

### 2.3 Aspect Attention Layer

In this layer, we implement the idea of discovering domain-specific aspects for cross-target stance inference. In particular, the key observation we make is that the domain aspects that reflect users' major concerns are usually the core of understanding their stances, and could be mentioned by multiple users in a discussion. For example, we find that many users in our corpus mention the aspect "reef" to express their concerns about the impact of a mining project on the Great Barrier Reef. Based on this observation, the perception of the domain aspects can be boiled down to finding the sentence parts that not only carry the core idea of a stance-bearing sentence but also tend to be recurring in the corpus.

First, to capture the recurrences of the domain aspects, a simple way is to make every input sentence be consumed by this layer (see Figure 1), so that the layer parameters are shared across the corpus for being stimulated by all appearances of the domain aspects.

Then, we utilize self-attention to signal the core parts of a stance-bearing sentence. Self-attention is an attention mechanism for selecting specific parts of a sequence by relating its elements at different positions (Vaswani et al., 2017; Cheng et al., 2016). In our case, the self-attention process is based on the assumption that the core parts of a sentence are those that are compatible with the semantics of the entire sentence. To this end, we introduce a compatibility function to score the semantic compatibility between the encoded se-

quence $\boldsymbol{H}^P$ and each of its hidden states $\boldsymbol{h}^P$:

$$c_i = \boldsymbol{w}_2^\top \sigma(\boldsymbol{W}_1 \boldsymbol{h}_i^P + \boldsymbol{b}_1) + b_2 \quad (3)$$

where $\boldsymbol{W}_1 \in \mathbb{R}^{d \times 2h}$, $\boldsymbol{w}_2 \in \mathbb{R}^d$, $\boldsymbol{b}_1 \in \mathbb{R}^d$, and $b_2 \in \mathbb{R}$ are trainable parameters, and $\sigma$ is the activation function. Note that all the above parameters are shared by every hidden state in $\boldsymbol{H}^P$. Next, we compute the attention weight $a_i$ for each $\boldsymbol{h}_i^P$ based on its compatibility score via softmax operation:

$$a_i = \frac{\exp(c_i)}{\sum_{j=1}^{|P|} \exp(c_j)} \quad (4)$$

Finally, we can obtain the domain aspect encoded representation based on the attention weights:

$$\boldsymbol{A}^P = \sum_{i=1}^{|P|} a_i \boldsymbol{h}_i^P \quad (5)$$

where $\boldsymbol{A}^P \in \mathbb{R}^{2h}$ is the domain aspect encoding for sentence $P$ and also the output of this layer.

### 2.4 Prediction Layer

We predict the stance label of the sentence based on its domain aspect encoding:

$$\hat{\boldsymbol{y}} = \text{softmax}(\text{MLP}(\boldsymbol{A}^P)) \quad (6)$$

where we use a multilayer perceptron (MLP) to consume the domain aspect encoding $\boldsymbol{A}^P$ and apply the softmax to get the predicted probability for each of the $C$ classes, $\hat{\boldsymbol{y}} = \{y_1, ..., y_C\}$.

### 2.5 Model Training

For model training, we use multi-class cross-entropy loss,

$$\mathcal{J}(\theta) = -\sum_i^N \sum_j^C y_j^{(i)} \log \hat{y}_j^{(i)} + \lambda \|\Theta\| \quad (7)$$

where $N$ is the size of training set. $y$ is the ground-truth label indicator for each class, and $\hat{y}$ is the predicted probability. $\lambda$ is the coefficient for $L_2$-regularization. $\Theta$ denotes the set of all trainable parameters in our model.

## 3 Experiments

This section reports the results of quantitative and qualitative evaluations of the proposed model.

| Target | %Favor | %Against | %Neither | #Total |
|---|---|---|---|---|
| CC | 59.4 | 4.6 | 36.0 | 564 |
| FM | 28.2 | 53.8 | 18.0 | 949 |
| HC | 16.6 | 57.4 | 26.0 | 984 |
| LA | 17.9 | 58.3 | 23.8 | 933 |
| DT | 20.9 | 42.3 | 36.8 | 707 |

Table 2: SemEval-2016 Task 6 Tweet Stance Detection dataset used in our evaluation.

### 3.1 Datasets

**SemEval-2016**: the first dataset is from SemEval-2016 Task 6 on Twitter stance detection, which contains stance-bearing tweets on different targets. We use the following five targets for our experiments: *Climate Change is Concern* (CC), *Feminist Movement* (FM), *Hillary Clinton* (HC), *Legalization of Abortion* (LA), and *Donald Trump* (DT). The class labels are *favor*, *against*, and *neither*, and their distributions are shown in Table 2.

**Tweets on an Australian mining project (AM):** the second is our collection of tweets on a mining project in Australia obtained using Twitter API. It includes 220,067 tweets posted from January 2016 to June 2017 that contain the project name in the text. We remove all URL-only tweets and duplicate tweets, and obtain a set of 40,852 (unlabeled) tweets. Due to the lack of annotation, this dataset is only used for our qualitative evaluation.

To align with our scenario, the above targets can be categorized into three different domains: Women's Rights (FM, LA), American Politics (HC, DT), and Environments (CC, AM).

### 3.2 Metric

We use *F1-score* to measure the classification performance. Due to the imbalanced class distributions of the SemEval dataset, we compute both micro-averaged (large classes dominate) and macro-averaged (small classes dominate) F1-scores (Manning et al., 2008), and use their average as the metric, i.e., $F = \frac{1}{2}(F_{\text{micro}} + F_{\text{macro}})$.

To evaluate the effectiveness of target adaptation, we use the metric *transfer ratio* (Glorot et al., 2011) to compare the cross-target and in-target performance of a model: $\mathcal{Q} = \frac{F(S,D)}{F_b(D,D)}$, where $F(S, D)$ is the cross-target F1-score of a model trained on the source target $S$ and tested on the destination target $D$, and $F_b(D, D)$ is the in-target F1-score of a *baseline* model trained and tested on the same target $D$, which serves as the performance calibration for target adaptation.

### 3.3 Training setup

The word embeddings are initialized with the pre-trained 200d GloVe word vectors on the 27B Twitter corpus (Pennington et al., 2014), and fixed during training. The model is trained (90%) and validated (10%) on a source target, and tested on a destination target. The following model settings are selected based on a small grid search on the validation set: the LSTM hidden size of 60, the MLP layer size of 60, and dropout 0.1. The $L$2-regularization coefficient $\lambda$ in the loss is 0.01. ADAM (Kingma and Ba, 2014) is used as the optimizer, with a learning rate of $10^{-3}$. Stratified 10-fold cross-validation is conducted to produce averaged results.

### 3.4 Classification Performance

This section reports the results of our model and two baseline approaches on cross-target stance classification.

**BiLSTM**: this is a base model for our task. It has two BiLSTMs for encoding the sentence and target separately. Then, the concatenation of the resulting encodings is fed into the final Prediction Layer to generate predicted stance labels. In our evaluation, this model is treated as the baseline model for deriving the in-target performance calibration $F_b(D, D)$.

**BiCondLSTM** (Augenstein et al., 2016) (denoted as **BiCond** hereafter): this is the best system in SemEval-2016 Task 6. It utilizes the conditional encoding to learn a target-dependent representation for the input sentence. The conditional encoding is realized in the same way as the Context Encoding Layer does in our model, namely by using the hidden states of the target-encoding BiLSTM to initialize the sentence-encoding BiLSTM.

Table 3 shows the results (in-target and cross-target) on the two domains: Women's Rights and American Politics. First, it is observed that Bi-Cond outperforms BiLSTM over all target configurations, suggesting that, compared to simple concatenation, the conditional encoding of the target information could be more helpful to capture the dependency of the sentence on the target.

Second, our model is shown to achieve better results than the two baselines in almost all cases (only slightly worse than BiCond on LA under the in-target setting, and the difference is not statistically significant), which implies that the aspect attention mechanism adopted in our model could benefit target-level generalization while it does not hurt the in-target performance. Moreover, by comparing the performance of our model under different target configurations, we see that the improvements brought by our model are more significant on the cross-target task than they are on the in-target task, with an average improvement of 6.6% (cross-target) vs. 3.0% (in-target) over BiCond in F1-score, which demonstrates a greater advantage of our model in the cross-target task.

Finally, according to the transfer ratio results, the general drop from the in-target to cross-target performance (26% averaged over all cases) could imply that while the target-independent information (i.e., the domain-specific aspects) is shown to benefit generalization, it could be important to also consider the information that is specific to the destination target for model building (which has not yet been explored in this work).

| Domain | Target | BiLSTM | BiCond | CrossNet |
|---|---|---|---|---|
| **In-target performance (F1-score)** | | | | |
| Women's Rights | FM | 51.4(2.6) | 53.7(3.9) | 55.9(4.6)[†] |
| | LA | 59.4(6.4) | 61.9(6.3) | 61.8(4.7) |
| American Politics | HC | 55.5(4.7) | 56.0(3.1) | 60.0(3.3)[†‡] |
| | DT | 57.9(6.0) | 59.6(5.8) | 60.2(5.1)[†‡] |
| **Cross-target performance (F1-score)** | | | | |
| Women's Rights | FM→LA | 40.1(2.3) | 40.3(2.2) | 44.2(1.3)[†‡] |
| | LA→FM | 37.9(2.9) | 39.2(1.5) | 43.1(1.3)[†‡] |
| American Politics | HC→DT | 43.3(2.4) | 44.2(3.2) | 46.1(3.7)[†] |
| | DT→HC | 40.1(2.6) | 40.8(2.2) | 41.8(3.2) |
| **Transfer ratio** | | | | |
| Women's Rights | FM→LA | 0.67 | 0.67 | 0.74 |
| | LA→FM | 0.74 | 0.76 | 0.83 |
| American Politics | HC→DT | 0.75 | 0.76 | 0.79 |
| | DT→HC | 0.72 | 0.73 | 0.75 |

(Standard deviation in parentheses)
[†]([‡]) Improvements over BiLSTM (BiCond) at $p < .05$ with paired t-test

Table 3: Classification performance of our model and other baselines on 4 targets: *Feminist Movement* (FM), *Hillary Clinton* (HC), *Legalization of Abortion* (LA), and *Donald Trump* (DT).

### 3.5 Visualization of Attention

To show that our model can select sentence parts that are related to domain aspects, we visualize the self-attention results on some tweet examples that are correctly classified by our model in Table 4.

We can see that the most highlighted parts in each example are relevant to the respective domain. For example, "feminist", "rights", and "equality" are commonly used when talking about women's rights, and "president" and "dreams" of-ten appear in text about politics. It is also interesting to note that words that are specific to the destination target may not be captured by the model learned from the source target, such as "abortion" in sentence 1 and "trumps" in sentence 3. This makes sense because those words are rare in the source target corpus and thus not well noticed by the model.

| ID | Target | Tweet |
|---|---|---|
| 1 | FM→LA | Abortion has nothing to do with feminism. Its about the BABYS body, not yours. (A) |
| 2 | LA→FM | All humans, male and female, should have equal political, economic and social rights. Equality. (F) |
| 3 | HC→DT | Trumps presidential dreams r about as real as KimJonguns unicorns. (A) |
| 4 | DT→HC | Maybe a woman should be President. (F) |
| 5 | CC→AM | [N] still will destroy the reef. It is criminal that QLD federal govts are promoting it. (A) |
| 6 | CC→AM | [N], who are trying to change our laws, 'forgot' to tell us about their CEO's environmental disaster! (A) |

Table 4: The heatmap of the attention weights assigned by the Aspect Attention Layer to the tweets with stance labels *favor* (F) and *against* (A). "[N]" denotes the mining project's name of interest.

| Domain | Sample |
|---|---|
| Women's Rights (FM↔LA) | feminist, victim, fight, equality, children, women, mom, great, love, fight, beautiful, girl, housewife, wrong, fear, safe, good, life, crime |
| American Politics (HC↔DT) | american, winning, welfare, obama, society, wall, crazy, money, republican, nation, great, vote, amazing, stupid, justice, government, future |
| Environments (CC→AM) | environment, reverse, needs, bad, weather, solution, important, rain, earth, generation, humans, power, heat, future, strategy, together, idea |

Table 5: Samples of the learned domain aspects.

Finally, for our project, we can see from the last two sentences that the model learned from *climate change is concern* is able to concentrate on words that are central to understanding the authors' stances on the approval of the mining project, such as "reef", "destroy", "environmental", and "disaster". Overall, the above visualization demonstrates that our model could benefit stance inference across related targets through capturing domain-specific information.

### 3.6 Learned Domain-Specific Aspects

Finally, it is also possible to show the learned domain aspects by extracting all sentence parts in a corpus that are highly attended by our model. Table 5 presents a number of samples from the intersections between the sets of highly-attended words on the respective targets in the three domains. Again, we see that these highly-attended words are specific to the respective domains. We also notice that besides the domain-aspect words, our model can find words that carry sentiments as well, such as "great", "crazy", and "beautiful", which contribute to stance prediction.

## 4 Conclusion and Future Work

In this work, we study cross-target stance classification and propose a novel self-attention neural model that can extract target-independent information for model generalization. Experimental results show that the proposed model can perceive high-level domain-specific information in a sentence and achieves superior results over a number of baselines in certain domains. In the future, there are several ways of extending our model.

First, selecting the effective source targets to generalize from is crucial for achieving satisfying results on the destination targets. One possibility could be to learn certain correlations between target closeness and generalization performance, which could further be used for guiding the target selection process. Second, our current model for identifying users' stances on mining projects only generalizes from one source target (i.e., *Climate Change is Concern*). However, a mining project in general could affect other aspects of our society such as community and economics. It could be useful to also consider other related sources for knowledge transfer. Finally, it would be interesting to evaluate our model in a multilingual scenario (Taulé et al., 2017), in order to examine its generalization ability (whether it can attend to useful domain-specific information in a new language) and multilingual scope.

## Acknowledgments

We thank all anonymous reviewers for their valuable comments. We would also like to thank Keith Vander Linden for his helpful comments on drafts of this paper.